\documentclass[11pt]{article}
\usepackage{acl2016}
\usepackage{times}
\usepackage{amssymb,latexsym,pifont}
\usepackage{graphicx}

\usepackage[utf8]{inputenc}

\usepackage{enumitem}
\usepackage{multirow}
\usepackage[normalem]{ulem}
\usepackage{todonotes}

\makeatletter
\newcommand{\@BIBLABEL}{\@emptybiblabel}
\newcommand{\@emptybiblabel}[1]{}
\makeatother
\usepackage{hyperref}

\definecolor{darkred}{rgb}{0.5, 0.0, 0.0}

\definecolor{darkblue}{rgb}{0.0, 0.0, 0.5}

\definecolor{darkgreen}{rgb}{0.0, 0.5, 0.0}

\def\miss#1{\textcolor{blue}{#1}}
\def\spfl#1{\textcolor{red}{#1}}
\def\rep#1{\textcolor{purple}{#1}}
\def\disfluent#1{\textcolor{brown}{#1}}

\newcommand\Tstrut{\rule{0pt}{2.3ex}}       
\newcommand\Bstrut{\rule[-0.9ex]{0pt}{0pt}} 

\aclfinalcopy 
\def\aclpaperid{430} 

\title{Sequence-to-Sequence Generation for Spoken Dialogue via Deep Syntax Trees and Strings}

\author{Ondřej Dušek \and Filip Jurčíček \\
Charles University in Prague, Faculty of Mathematics and Physics \\
Institute of Formal and Applied Linguistics \\
Malostranské náměstí 25, CZ-11800 Prague, Czech Republic \\
\url{{odusek,jurcicek}@ufal.mff.cuni.cz}
}

\date{}

\begin{document}

\maketitle

\begin{abstract}
We present a natural language generator based on the sequence-to-sequence approach that can be trained to produce natural language strings as well as deep syntax dependency trees from input dialogue acts, and we use it to directly compare two-step generation with separate sentence planning and surface realization stages to a joint, one-step approach.

We were able to train both setups successfully using very little training data. The joint setup offers better performance, surpassing state-of-the-art with regards to $n$-gram-based scores while providing more relevant outputs. 
\end{abstract}


\section{Introduction}

In spoken dialogue systems (SDS), the task of natural language generation (NLG) is to convert a meaning representation (MR) produced by the dialogue manager into one or more sentences in a natural language. It is traditionally divided into two subtasks: \emph{sentence planning}, which decides on the overall sentence structure,
and \emph{surface realization}, determining the exact word forms and linearizing the structure into a string \cite{reiter_building_2000}.
While some generators keep this division and use a two-step pipeline \cite{walker_spot:_2001,rieser_optimising_2010,dethlefs_conditional_2013}, others apply a joint model for both tasks \cite{wong_generation_2007,konstas_global_2013}.


We present a new, conceptually simple NLG system for SDS
that is able to operate in both modes: it either produces natural language strings or generates deep syntax dependency trees, which are subsequently processed by an external surface realizer \cite{dusek_new_2015}.
This allows us to show a direct comparison of two-step generation, where sentence planning and surface realization are separated, with a joint, one-step approach.

Our generator is based on the sequence-to-sequence (seq2seq) generation technique \cite{cho_learning_2014,sutskever_sequence_2014},
combined with beam search and an $n$-best list reranker to suppress irrelevant information in the outputs.
Unlike most previous NLG systems for SDS (e.g., \cite{stent_trainable_2004,raux_lets_2005,mairesse_phrase-based_2010}), it is trainable from unaligned pairs of MR and sentences alone.
We experiment with using much less training data than recent
systems based on recurrent neural networks (RNN) \cite{wen_semantically_2015,mei_what_2015}, and we find that our generator learns successfully to produce both strings and deep syntax trees on the BAGEL restaurant information dataset \cite{mairesse_phrase-based_2010}. It is able to surpass $n$-gram-based scores achieved previously by \newcite{dusek_training_2015}, offering a simpler setup and more relevant outputs.

We introduce the generation setting in Section~\ref{sec:task} and describe our generator architecture in Section~\ref{sec:model}. Section~\ref{sec:experiments} details our experiments, Section~\ref{sec:results} analyzes the results. We summarize related work in Section~\ref{sec:related} and offer conclusions in Section~\ref{sec:concl}.

\section{Generator Setting}\label{sec:task}

\begin{figure}
\begin{center}
\includegraphics[height=5cm]{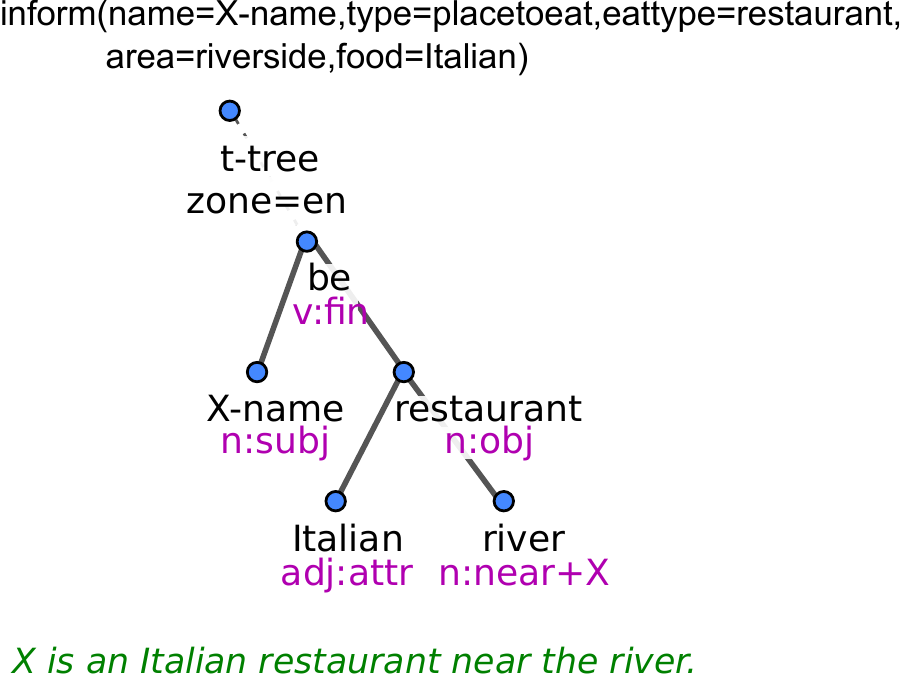}
\end{center}\vspace{-5mm}
\caption{Example DA (top) with the corresponding deep syntax tree (middle) and natural language string (bottom)}\label{fig:sentplan}
\vspace{-4mm}
\end{figure}

The input to our generator are \emph{dialogue acts} (DA) \cite{young_hidden_2010} 
representing an action, such as \emph{inform} or \emph{request}, along with one or more attributes (\emph{slots}) and their values.
Our generator operates in two modes, producing either deep syntax trees \cite{dusek_formemes_2012} or natural language strings (see Fig.~\ref{fig:sentplan}).
The first mode corresponds to the sentence planning NLG stage 
as it decides the syntactic shape of the output sentence; the resulting deep syntax tree involves content words (lemmas) and their syntactic form (formemes, purple in Fig.~\ref{fig:sentplan}).
The trees are linearized to strings using a surface realizer from the TectoMT translation system \cite{dusek_new_2015}.
The second generator mode joins sentence planning and surface realization into one step, producing natural language sentences directly.

Both modes offer their advantages: The two-step mode simplifies generation by abstracting away from complex surface syntax and morphology, which can be handled by a handcrafted, domain-independent module to ensure grammatical correctness at all times 
\cite{dusek_training_2015}, and the joint mode does not need to model structure explicitly and avoids accumulating errors along the pipeline \cite{konstas_global_2013}.


\section{The Seq2seq Generation Model}\label{sec:model}
\vspace{-1mm}

Our generator is based on the seq2seq approach \cite{cho_learning_2014,sutskever_sequence_2014}, a type of an encoder-decoder RNN architecture operating on variable-length sequences of tokens.
We address the necessary conversion of input DA and output trees/sentences into sequences in Section~\ref{sec:embeddings} and then
describe the main seq2seq component 
in Section~\ref{sec:seq2seq-model}. It is supplemented by a reranker
, as explained in Section~\ref{sec:reranking-classifier}.

\subsection{Sequence Representation of DA, Trees, and Sentences}\label{sec:embeddings}
\vspace{-1mm}

\begin{figure*}[tb]
\begin{center}
\includegraphics[height=0.7cm]{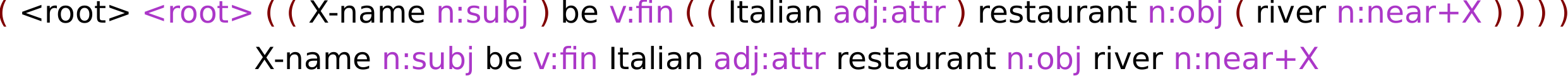}
\end{center}\vspace{-5mm}
\caption{Trees encoded as sequences for the seq2seq generator (top) and the reranker (bottom)}
\label{fig:tree-emb}
\end{figure*}

We represent DA, deep syntax trees, and sentences as sequences of tokens to enable their usage in the sequence-based RNN components of our generator (see Sections~\ref{sec:seq2seq-model} and~\ref{sec:reranking-classifier}).
Each token is represented by its embedding – a vector of floating-point numbers \cite{bengio_neural_2003}. 

To form a sequence representation of a DA, we create a triple of the structure “DA type, slot, value” for each slot in the DA
and concatenate the triples (see Fig.~\ref{fig:seq2seq}).
The deep syntax tree output from the seq2seq generator is represented in a bracketed notation similar to the one used by \newcite[see Fig.~\ref{fig:tree-emb}]{vinyals_grammar_2015}.
The inputs to the reranker are always a sequence of tokens; structure is disregarded in trees, resulting in a list of lemma-formeme pairs (see Fig.~\ref{fig:tree-emb}).



\vspace{-0.5mm}
\subsection{Seq2seq Generator}\label{sec:seq2seq-model}

\begin{figure*}[tb]\vspace{-1mm}
\begin{center}
\vspace{2mm}
\includegraphics[height=2.5cm]{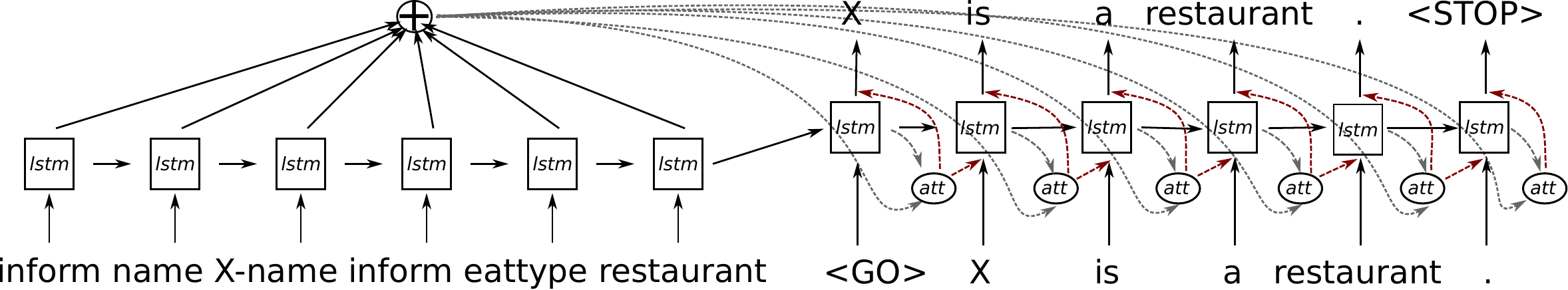}
\end{center}\vspace{-5mm}
\caption{Seq2seq generator with attention}\label{fig:seq2seq}\vspace{-4mm}
\end{figure*}

\vspace{-0.5mm}
Our seq2seq generator with attention \cite[see Fig.~\ref{fig:seq2seq}]{bahdanau_neural_2014}\footnote{We use the implementation in the TensorFlow framework \cite{tensorflow2015-whitepaper}.} 
starts with the encoder stage, which uses an RNN to encode an input sequence $\mathbf{x} = \{x_1,\dots,x_n\}$ into a sequence of encoder outputs and hidden states $\mathbf{h} = \{h_1,\dots,h_n\}$, where $h_t = \mbox{lstm}(x_t,h_{t-1})$, a non-linear function represented by the long-short-term memory (LSTM) cell \cite{graves_generating_2013}.

The decoder stage then uses the hidden states
to generate a sequence $\mathbf{y} = \{y_1,\dots,y_m\}$ with a second LSTM-based RNN. The probability of each output token is defined as:
\vspace{-2mm}
$$p(y_t|y_1,\dots,y_{t-1},\mathbf{x}) = \mbox{softmax}((s_t\circ c_t)W_Y)$$

\vspace{-2mm}\noindent
Here, $s_t$ is the decoder state where $s_0 = h_n$ and $s_t = \mbox{lstm}((y_{t-1}\circ c_t)W_S,s_{t-1})$, i.e., the decoder is initialized by the last hidden state and uses the previous output token at each step.
$W_Y$ and $W_S$ are learned linear projection matrices and “$\circ$” denotes concatenation.
$c_t$ is the \emph{context vector} – a weighted sum of the encoder hidden states $c_t = \sum_{i=1}^n \alpha_{ti} h_i$, where $\alpha_{ti}$ corresponds to an \emph{alignment model}, represented by a feed-forward network with a single $\tanh$ hidden layer.

On top of this basic seq2seq model, we implemented a simple beam search for decoding
\cite{sutskever_sequence_2014,bahdanau_neural_2014}.
It proceeds left-to-right and keeps track of log probabilities of top $n$ possible output sequences,
expanding them one token at a time.

\subsection{Reranker}\label{sec:reranking-classifier}

\begin{figure}[tb]
\begin{center}
\includegraphics[height=3.9cm]{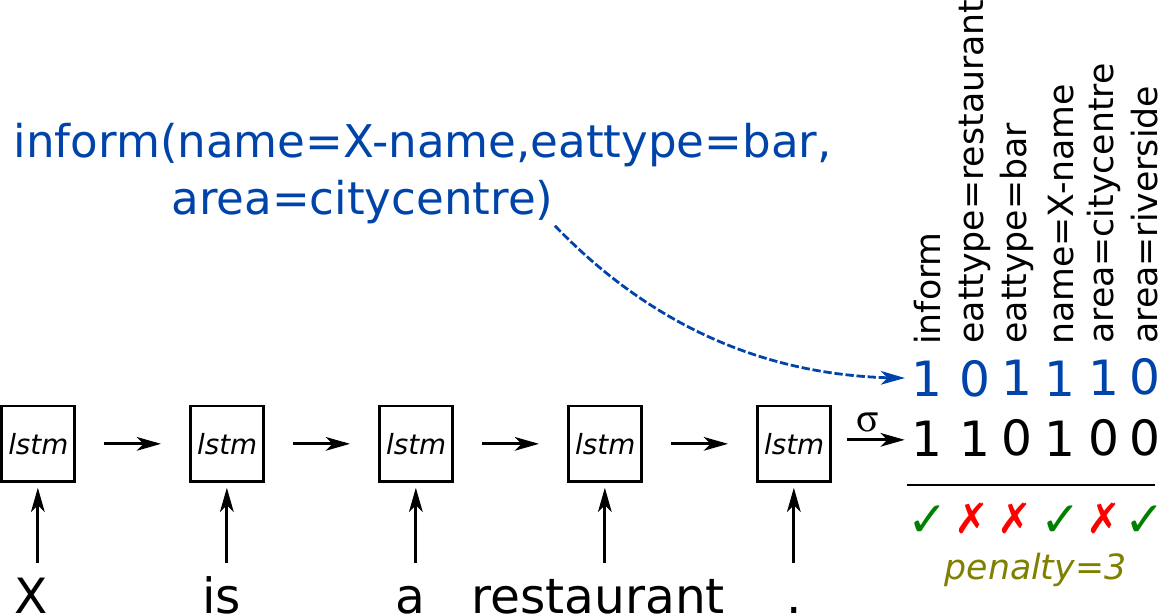}
\end{center}\vspace{-5mm}
\caption{The reranker}\label{fig:rerank}\vspace{-5mm}
\end{figure}
\vspace{-1mm}
To ensure that the output trees/strings correspond semantically to the input DA, we implemented a classifier to rerank the $n$-best beam search outputs and penalize those 
missing required information and/or adding irrelevant one.
Similarly to \newcite{wen_stochastic_2015}, the classifier provides a binary decision for an output tree/string on the presence of all dialogue act types and slot-value combinations seen in the training data, producing a 1-hot vector. The input DA is converted to a similar 1-hot vector
and the reranking penalty of the sentence is the Hamming distance between the two vectors (see Fig.~\ref{fig:rerank}).
Weighted penalties for all sentences are subtracted from their $n$-best list log probabilities.


We employ a similar architecture for the classifier as in our seq2seq generator encoder (see Section~\ref{sec:seq2seq-model}), with an RNN encoder operating on the output trees/strings and a single logistic layer for classification over the last encoder hidden state.
Given an output sequence representing a string or a tree $\mathbf{y} = \{y_1,\dots,y_n\}$ (cf.~Section~\ref{sec:embeddings}), the encoder again produces a sequence of hidden states $\mathbf{h} = \{h_1,\dots,h_n\}$ where $h_t = \mbox{lstm}(y_t,h_{t-1})$.
The output binary vector $o$ is computed as:
\vspace{-2mm}
$$o_i= \mbox{sigmoid}((h_n \cdot W_R + b)_i)$$

\vspace{-2mm}\noindent
Here, $W_R$ is a learned projection matrix and $b$ is a corresponding bias term.

\section{Experiments}\label{sec:experiments}

We perform our experiments on the 
BAGEL data set of \newcite{mairesse_phrase-based_2010}, which contains 202 DA from the restaurant information domain with two natural language paraphrases each, describing restaurant locations, price ranges, food types etc.
Some properties such as restaurant names or phone numbers are delexicalized (replaced with “X” symbols) to avoid data sparsity.\footnote{We adopt the delexicalization scenario used by \newcite{mairesse_phrase-based_2010} and \newcite{dusek_training_2015}.}
Unlike \newcite{mairesse_phrase-based_2010}, we do not use manually annotated alignment of slots and values in the input DA to target words and phrases and let the generator learn it from data, which simplifies training data preparation but makes our task harder. 
We lowercase the data and treat plural \mbox{\emph{-s}} as separate tokens for generating into strings, and
we apply automatic analysis from the Treex NLP toolkit \cite{popel_tectomt:_2010} to obtain deep syntax trees for training tree-based generator setups.\footnote{The input vocabulary size is around 45 (DA types, slots, and values added up) and output vocabulary sizes are around 170 for string generation and 180 for tree generation (45 formemes and 135 lemmas).}
Same as \newcite{mairesse_phrase-based_2010}, we apply 10-fold cross-validation, with 181 training DA and 21 testing DA. In addition, we reserve 10 DA from the training set for validation.\footnote{We treat the two paraphrases for the same DA as separate instances in the training set but use them together as two references to measure BLEU and NIST scores \cite{papineni_bleu:_2002,doddington_automatic_2002} on the validation and test sets.}

To train our seq2seq generator, we use the Adam optimizer \cite{kingma_adam:_2015} to minimize unweighted sequence cross-entropy.\footnote{Based on a few preliminary experiments, the learning rate is set to 0.001, embedding size 50, LSTM cell size 128, and batch size 20. Reranking penalty for decoding is 100.} We perform 10 runs with different random initialization of the network and up to 1,000 passes over the training data,\footnote{Training is terminated early if the top 10 so far achieved validation BLEU scores do not change for 100 passes.} validating after each pass and selecting the parameters that yield the highest BLEU score on the validation set.
Neither beam search nor the reranker are used for validation.

We use the Adam optimizer minimizing cross-entropy to train the reranker as well.\footnote{We use the same settings as with the seq2seq generator.} We perform a single run of up to 100 passes over the data, and we also validate after each pass and select the parameters giving minimal Hamming distance on both validation and training set.\footnote{The validation set is given 10 times more importance.}

\section{Results}\label{sec:results}

\begin{table}[tb]
\addtolength{\tabcolsep}{-1mm}
\begin{tabular}{lccc}
\bf Setup                & \bf BLEU & \bf NIST   & \bf ERR\hspace{-1mm} \\\hline
\newcite{mairesse_phrase-based_2010}$^\ast$ & $\sim$67 & -      & \phantom{0}0 \\
\newcite{dusek_training_2015}\hspace{-1mm}  & 59.89    & 5.231  & 30 \\\hline
%
%
%
Greedy with trees        	    & 55.29 & 5.144 & 20 \\ 
+ Beam search (b.~size 100)         & 58.59 & 5.293 & 28 \\ 
+ Reranker (beam size 5)            & 60.77 & 5.487 & 24 \\ 
\phantom{+ Reranker} (beam size 10) & 60.93 & 5.510 & 25 \\ 
\phantom{+ Reranker} (beam size 100)& 60.44 & 5.514 & 19 \\ 
Greedy into strings                 & 52.54 & 5.052 & 37 \\ 
+ Beam search (b.~size 100)         & 55.84 & 5.228 & 32 \\ 
+ Reranker (beam size 5)            & 61.18 & 5.507 & 27 \\ 
\phantom{+ Reranker} (beam size 10) & 62.40 & 5.614 & 21 \\ 
\phantom{+ Reranker} (beam size 100)& 62.76 & 5.669 & 19 \\ 
\end{tabular}
\addtolength{\tabcolsep}{2mm}\vspace{-4mm}
\caption{Results on the BAGEL data set}\label{tab:results}

\small\smallskip
NIST, BLEU, and semantic errors in a sample of the output.

\smallskip\noindent
$^\ast$\newcite{mairesse_phrase-based_2010} use manual alignments in their work, so their result is not directly comparable to ours. The zero semantic error is implied by the manual alignments and the architecture of their system.
\vspace{-5mm}
\end{table}

\begin{table*}[tb]
\begin{center}\footnotesize
\begin{tabular}{ll}
Input DA &    inform(name=X-name, type=placetoeat, eattype=restaurant, area=citycentre, near=X-near, \\
   &  \phantom{inform(}food="Chinese takeaway", food=Japanese)\\
Reference &   X is a Chinese takeaway and Japanese restaurant in the city centre near X.\\
Greedy with trees &     X is a restaurant offering chinese takeaway in the centre of town near X. [\miss{Japanese}]\\
+ Beam search &   X is a restaurant \disfluent{and} japanese food \disfluent{and} chinese takeaway.\\
+ Reranker &  X is a restaurant serving japanese food in the centre of the city that offers chinese takeaway.\\
Greedy into strings &     X is a restaurant offering \spfl{italian} and \spfl{indian} takeaway in the city centre area near X. \hbox to 0cm{[\miss{Japanese}, \miss{Chinese}]}\\
+ Beam search &   X is a restaurant that serves \spfl{fusion} chinese takeaway in the \spfl{riverside} area near X. \hbox to 0cm{[\miss{Japanese}, \miss{citycentre}]}\\
+ Reranker &  X is a japanese restaurant in the city centre near X providing chinese food. [\miss{takeaway}]\Bstrut \\\hline

Input DA &    inform(name=X-name, type=placetoeat, eattype=restaurant, area=riverside, food=French)\Tstrut\\
Reference &   X is a French restaurant on the riverside.\\
Greedy with trees &     X is a restaurant providing french and \spfl{continental} \disfluent{and} by the river.\\
+ Beam search &   X is a restaurant that serves french \spfl{takeaway}. [\miss{riverside}]\\
+ Reranker &  X is a french restaurant in the riverside area.\\
Greedy into strings &     X is a restaurant in the riverside that serves \spfl{italian} food. [\miss{French}]\\
+ Beam search &   X is a restaurant in the riverside that serves \spfl{italian} food. [\miss{French}]\\
+ Reranker &  X is a restaurant in the riverside area that serves french food.\Bstrut \\\hline

Input DA &     inform(name=X-name, type=placetoeat, eattype=restaurant, near=X-near, food=Continental, \hbox to 0.95cm{food=French)}\Tstrut\\
Reference &   X is a French and Continental restaurant near X.\\
Greedy with trees &     X is a french restaurant that serves \rep{french} food \disfluent{and} near X. [\miss{Continental}]\\
+ Beam search &   X is a french restaurant that serves \rep{french} food \disfluent{and} near X. [\miss{Continental}]\\
+ Reranker &  X is a restaurant serving french and continental food near X.\\
Greedy into strings &     X is a french and continental style restaurant near X.\\
+ Beam search &   X is a french and continental style restaurant near X.\\
+ Reranker &  X is a restaurant providing french and continental food, near X.\\
\end{tabular}
\end{center}
\vspace{-2mm}
\caption{Example outputs of different generator setups (beam size 100 is used). Errors are marked in color (\miss{missing}, \spfl{superfluous}, \rep{repeated} information, \disfluent{disfluency}).}\label{tab:outputs}
\vspace{-2mm}
\end{table*}

\vspace{-1mm}
The results of our experiments and a comparison to previous works on this dataset are shown in Table~\ref{tab:results}.
We include BLEU and NIST scores and the number of semantic errors (incorrect, missing, and repeated information), which we assessed manually on a sample of 42 output sentences (outputs of two randomly selected cross-validation runs).

The outputs of direct string generation show that the models learn to produce fluent sentences in the domain style;\footnote{The average sentence length is around 13 tokens.} incoherent sentences
are rare, but semantic errors are very frequent in the greedy search.
Most errors involve confusion of semantically close items, e.g., \emph{Italian} instead of \emph{French} or \emph{riverside area} instead of \emph{city centre} (see Table~\ref{tab:outputs}); items occurring more frequently are preferred regardless of their relevance.
The beam search brings a BLEU improvement but keeps most semantic errors in place.
The reranker is able to reduce the number of semantic errors while increasing automatic scores considerably.
Using a larger beam increases the effect of the reranker as expected, resulting in slightly improved outputs.

Models generating deep syntax trees are also able to learn the domain style, and they have virtually no problems producing valid trees.\footnote{The generated sequences are longer, but have a very rigid structure, i.e., less uncertainty per generation step. The average output length is around 36 tokens in the generated sequence or 9 tree nodes; surface realizer outputs have a similar length as the sentences produced in direct string generation.}
The surface realizer works almost flawlessly on this limited domain 
\cite{dusek_training_2015}, 
leaving the seq2seq generator as the major error source.
The syntax-generating models tend to make different kinds of errors than the string-based models: Some outputs are valid trees but not entirely syntactically fluent; missing, incorrect, or repeated information is more frequent than a confusion of semantically similar items (see Table~\ref{tab:outputs}).
Semantic error rates of greedy and beam-search decoding are lower than for 
string-based
models, partly because confusion of two similar items counts as two errors. 
The beam search brings an increase in BLEU but also in the number of semantic errors. 
The reranker is able to reduce the number of errors and improve automatic scores slightly.
A larger beam leads to a small BLEU decrease even though the sentences contain less errors; here, NIST reflects the situation more accurately.

A comparison of the two approaches goes in favor of the joint setup: Without the reranker, models generating trees produce less semantic errors and gain higher BLEU/NIST scores. However, with the reranker, the string-based model is able to reduce the number of semantic errors while producing outputs significantly better in terms of BLEU/NIST.\footnote{The difference is statistically significant at 99\% level according to pairwise bootstrap resampling test \cite{koehn_statistical_2004}.}
In addition, the joint setup does not need an external surface realizer.
The best results of both setups surpass the best results on this dataset using training data without manual alignments \cite{dusek_training_2015} in both 
automatic metrics\footnote{The BLEU/NIST differences are statistically significant.} and the number of semantic errors.

\vspace{-1mm}
\section{Related Work}\label{sec:related}

\vspace{-1mm}
While most recent NLG systems attempt to learn generation from data, the choice of a particular approach – pipeline or joint – is often arbitrary and depends on system architecture or particular generation domain.
Works using the pipeline approach in SDS tend to focus on sentence planning, improving a handcrafted generator \cite{walker_spot:_2001,stent_trainable_2004,paiva_empirically-based_2005} or using perceptron-guided A* search \cite{dusek_training_2015}.
Generators taking the joint approach employ various methods, e.g., 
factored language models \cite{mairesse_phrase-based_2010}, inverted parsing 
\cite{wong_generation_2007,konstas_global_2013}, or a pipeline of discriminative classifiers \cite{angeli_simple_2010}.
Unlike most previous NLG systems, our generator is trainable from unaligned pairs of MR and sentences alone.

Recent RNN-based generators are most similar to our work.
\newcite{wen_stochastic_2015} combined two RNN with a convolutional network reranker; \newcite{wen_semantically_2015} later replaced basic sigmoid cells with an LSTM.
\newcite{mei_what_2015} present the only seq2seq-based NLG system known to us. 
We extend the previous works by generating deep syntax trees as well as strings and directly comparing pipeline and joint generation. 
In addition, we experiment with an order-of-magnitude smaller dataset
than other RNN-based systems.

\section{Conclusions and Future Work}\label{sec:concl}

\vspace{-1mm}
We have presented a direct comparison of two-step generation via deep syntax trees with a direct generation into strings, both using the same NLG system based on the seq2seq approach. While both approaches offer decent performance, their outputs are quite different. The results show the direct approach as more favorable, with significantly higher $n$-gram based scores and a similar number of semantic errors in the output.

We also showed that our generator can learn to produce meaningful utterances using a much smaller amount of training data than what is typically used for RNN-based approaches. The resulting models had virtually no problems with producing fluent, coherent sentences or with generating valid structure of bracketed deep syntax trees.
Our generator was able to surpass the best BLEU/NIST scores on the same dataset previously achieved by a perceptron-based generator of \newcite{dusek_training_2015} while reducing the amount of irrelevant information on the output.

Our generator is released on GitHub at the following URL:
\begin{center}\vspace{-1mm}
\url{https://github.com/UFAL-DSG/tgen}
\end{center}
We intend to apply it to other datasets for a broader comparison, and we plan further improvements, such as enhancing the reranker or including a bidirectional encoder \cite{bahdanau_neural_2014,mei_what_2015,jean_using_2015} and sequence level training \cite{ranzato_sequence_2015}.


\section*{Acknowledgments}

\vspace{-1mm}
This work was funded by the Ministry of Education, Youth and Sports of the Czech Republic
under the grant agreement LK11221 and core research funding, SVV project 260~333, and GAUK
grant 2058214 of Charles University in Prague. It used language resources stored and distributed by
the LINDAT/CLARIN project of the Ministry of Education, Youth and Sports of the Czech Republic (project LM2015071).
We thank our colleagues and the anonymous reviewers for helpful comments.

\bibliographystyle{acl2016}
\bibliography{paper}

\end{document}